\documentclass[letterpaper, 10 pt, conference]{ieeeconf}  

\IEEEoverridecommandlockouts                              

\usepackage{amsmath} 
\usepackage{amssymb}  
\usepackage{multirow}

\usepackage{graphicx}
\usepackage{caption}
\usepackage{subcaption}
\usepackage{hyperref}
\usepackage{gensymb} 
\usepackage{pifont}

\title{\LARGE \bf
Point Cloud Based Reinforcement Learning for Sim-to-Real and Partial Observability in Visual Navigation
}

\author{Kenzo Lobos-Tsunekawa$^{1}$ and Tatsuya Harada$^{12}$
\thanks{$^{1}$ is with the Graduate School of Information Science and Technology,
The University of Tokyo, Japan. Email: {\tt\small \{kenzolobos,harada\}@mi.t.u-tokyo.ac.jp}
        }%
\thanks{$^{2}$ is with the RIKEN Center for Advanced Intelligence Project (RIKEN
AIP), Tokyo, Japan. Email: \tt\small tatsuya.harada@riken.jp}%
}

\begin{document}

\maketitle
\thispagestyle{empty}
\pagestyle{empty}

\begin{abstract}

Reinforcement Learning (RL), among other learning-based methods, represents powerful tools to solve complex robotic tasks (e.g., actuation, manipulation, navigation, etc.), with the need for real-world data to train these systems as one of its most important limitations. The use of simulators is one way to address this issue, yet knowledge acquired in simulations does not work directly in the real-world, which is known as the sim-to-real transfer problem. While previous works focus on the nature of the images used as observations (e.g., textures and lighting), which has proven useful for a sim-to-sim transfer, they neglect other concerns regarding said observations, such as precise geometrical meanings, failing at robot-to-robot, and thus in sim-to-real transfers. We propose a method that learns on an observation space constructed by point clouds and environment randomization, generalizing among robots and simulators to achieve sim-to-real, while also addressing partial observability. We demonstrate the benefits of our methodology on the point goal navigation task, in which our method proves to be highly unaffected to unseen scenarios produced by robot-to-robot transfer, outperforms image-based baselines in robot-randomized experiments, and presents high performances in sim-to-sim conditions. Finally, we perform several experiments to validate the sim-to-real transfer to a physical domestic robot platform, confirming the out-of-the-box performance of our system.  

\end{abstract}

\section{INTRODUCTION}

Reinforcement Learning (RL) is a Machine Learning paradigm used to solve problems modeled as Markov Decision Process (MDP), which has presented important achievements in several areas, being board games \cite{Silver17}, videogames \cite{mnih2015} and robotic tasks \cite{ZhuMKLGFF17} among the most common ones.


As most complex robotic tasks (e.g., locomotion, navigation, manipulation, etc.) have a sequential nature and are difficult to model, RL becomes highly convenient, as it provides a framework which focuses on the desired behavior of the agent (i.e., reward function) rather than its implementation (i.e., explicit design of the policy), by learning a policy through interactions with the environment.

\begin{figure}
    \centering
    \vspace*{2.5mm}
    \begin{subfigure}[b]{0.2\textwidth}
        \centering
        \captionsetup{justification=centering}
        \includegraphics[width=\textwidth]{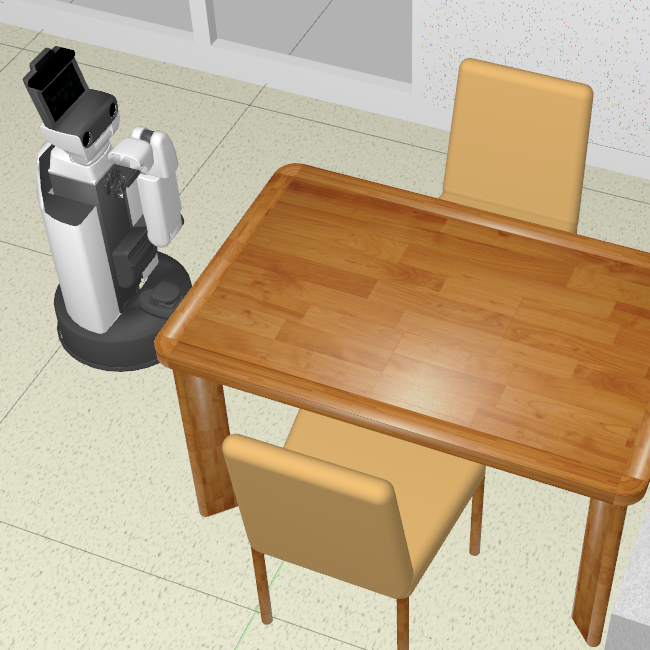}
        \caption{Simulated environment}
        \label{fig:hsr-simulated}
    \end{subfigure}
    \qquad
    \begin{subfigure}[b]{0.2\textwidth}
        \centering
        \captionsetup{justification=centering}
        \includegraphics[width=\textwidth]{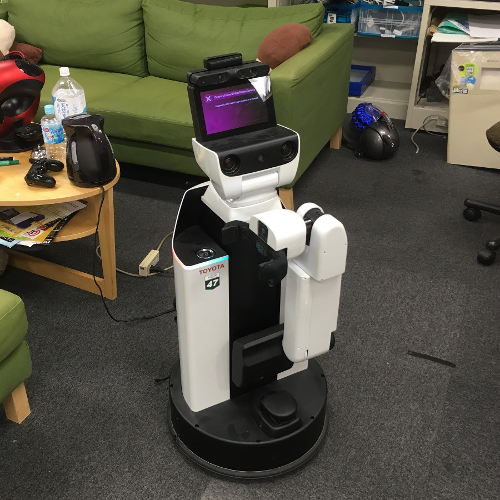}
        \caption{Real environment}
        \label{fig:hsr-real}
    \end{subfigure}
    \caption{Comparison between environments.}
    \vspace{-2.0em}
    \label{fig:sim2real}
\end{figure}

However, several difficulties limit the widespread use of RL and other learned-based methods in robots. Since RL approaches require an enormous amount of data/interactions for training, and the nature of RL is based on learning from previous mistakes, implementing the training procedure in the robot is impractical as it is both time-consuming and dangerous (or costly). For these reasons, most successful RL applications for robots rely on training policies on simulators and then deploying them on real robots (Figure \ref{fig:sim2real} shows examples of simulated and real environments). Doing so, however, rises a new problem. Since the environment in which the policy is evaluated is different from the environment it was trained on, and due to the sequential nature of MDPs, policy trajectories differ considerably, and in many cases,  renders the policies useless. This is a classic problem that is usually referred to as sim-to-real transfer or more generally as the reality gap, and it has been intensively addressed by different approaches. Some examples are: matching train/evaluation distributions \cite{gibonenv18}, \cite{Savva2019}, \cite{ZhangRAL19Goggles}, trying to consider random dynamics \cite{PengICRA2018}, and using some small real-world set of interactions to either adapt or retrain policies on the real robot \cite{rakelly19}. On another note, the most successful applications of RL involve tasks that are known to be fully observable \cite{Silver17}, \cite{mnih2015}. Since RL considers MDPs, when observations are used instead of states, performance is strongly affected. A common approach to address this issue is to use Recurrent Neural Networks (RNN) to integrate temporal information. While this approach attains good results in many tasks, it still has many limitations, including handling long sequences. To address these, external memories have been proposed \cite{chaplot2018active}, \cite{guss2019minerlcomp}, \cite{chaplot2020learning}. Although their use as observations is direct, they are usually task-specific (e.g., maps for navigation), may require information that is usually not available at evaluation time (e.g., ground-truth information), and are non-differentiable, so they can not be learned end-to-end.

While previous attempts to address sim-to-real transfer and partial observability have achieved varying levels of success, current state-of-the-art methods are not enough to seamlessly transfer policies trained in simulators to the real world. To achieve this goal, our main proposal is based on the use of point clouds to represent the state-space, instead of the most commonly used RGB or RGBD images in robotic tasks, whenever applicable. The reasoning behind this is that point clouds projected in the cartesian space become highly independent of the robots and its sensors, making it a canonical space suited for learning representations capable of sim-to-real transfers. Additionally, the nature of point clouds allows us to address up to some extent the partial observability problem, as we can explicitly exploit previous information by integrating it into the representation.

To test out these hypotheses, we consider a visual navigation task as our case of study, as it features a high-dimensional observation space, requires long temporal sequences, and has been intensively used as a testbed for sim-to-real approaches in the past \cite{gordonICCV19splitnet}, \cite{Savva2019}, \cite{gibonenv18}, \cite{JamesCVPR19}. 

Our experiments indicate that in simple settings, our approach achieves similar performance to the baselines, yet in more realistic conditions our method considerably outperforms the image-based baseline in all configurations. It is highly unaffected by unseen robot configurations (robot-to-robot transfers) whereas the baseline fails completely, presents higher performances when environment randomization is used during training, and obtains a better sim-to-sim  capability. Finally, during real-world experiments, only the proposed method attains satisfactory results with an out-of-the-box configuration, validating its sim-to-real performance.
 
The main contributions of this paper are: 1) we propose a method to achieve out-of-the-box sim-to-real, and validate it with real-world experiments, 2) we provide a strategy to explicitly address the limited observability proper of robotic tasks. 3) we present a point cloud network design that extracts multi-scale features, designed for its use with the large batch sizes used in recent popular RL algorithms, while taking into account the limitations of current GPUs.


\section{Related work}\label{sec:related}

\subsection{Visual Navigation}

Visual navigation encompasses a broad range of tasks (e.g., navigation towards given coordinates, objects, etc.), in which embodied agents (e.g., robots) must traverse through an environment to accomplish a certain objective \cite{Mishkin19}, \cite{Savva2019}.  These tasks have been thoroughly addressed by the robotics community, and classic methods usually split the problems in separate modules, one being mapping and one or more levels of planning algorithms \cite{Mishkin19}. However, their performance is conditioned on the availability of high-quality maps, and they perform poorly when used on unseen environments, due to a high number of environment-specific parameters.

Recently, great focus has been put into learning-based navigation systems \cite{Anderson18}, \cite{Mishkin19}, which aim to improve their overall performance and generalize to unseen environments. Although in the machine learning community there have been several works related to visual navigation in either game-like domains \cite{JaderbergMCSLSK16}, \cite{guss2019minerlcomp} or robotic environments \cite{LobosTsunekawa2018}, \cite{Levia19}, the robotics community has been reluctant to adopt these systems, believing classic methods to perform better for generic tasks \cite{kojima19}. In this context, \cite{Savva2019} shows that with sufficient training, RL-based methods can outperform classic methods on fair settings, and \cite{wijmans2019ddppo} shows that RL agents can solve visual navigation tasks with an almost perfect SPL score \cite{Anderson18} in simulations. We believe this to be the right direction, but there are still several limitations that need to be addressed, as we show in Section \ref{sec:results}.

\subsection{Sim-to-Real}

The true potential of RL lies in the applications it can solve in the real world. Consequently, a considerable amount of research has been dedicated into making RL work in real scenarios. Current RL methods require an vast amount of environment interactions to learn high-performance policies, so most successful implementations rely on first training the policy on simulated environments and then using several techniques to transfer these policies to the real world. Some strategies consist on adapting the policy to the real-world (e.g., retraining parts of the policy \cite{Zhang17f} and meta-RL \cite{rakelly19}), considering domain-randomization \cite{PengICRA2018} during training,
and attempting to match the training/test observations or embeddings (e.g., using segmented images \cite{LobosTsunekawa2018}, projecting real-world observations to the ones used in simulations \cite{gibonenv18}, \cite{ZhangRAL19Goggles}, \cite{JamesCVPR19}, and using photo-realistic simulators \cite{gibonenv18}, \cite{Savva2019}).

One of the cases with the best results in terms of sim-to-real is the case of navigation tasks using range lasers \cite{Tai2017}, since geometry and observations are similar between simulation and reality. However, these approaches usually use lasers as feature vectors \cite{Tai2017} and are only valid for tasks in which the 2D observations are enough. Following the idea of using depth sensors for a seamless transfer, the use of depth cameras is logical and have been used in several works \cite{Savva2019}, \cite{Levia19}. However, the nature of these sensors differs greatly from the real ones (e.g., reflections, invalid values, etc.), and if the simulated robot is not completely matched to the real one, the approach is prone to fail. To prevent this phenomenon, up to the finest details of the real robot and its sensors are attempted to be replicated in simulations.

Following this line, recent simulators aim for photo-realism by reconstructing real environments \cite{Savva2019}, \cite{gibonenv18}, \cite{Matterport3D}. Authors suggest that by training agents in these environments, a successful sim-to-real transfer can be obtained, and give as a partial proof that this approach attains good results on the related sim-to-sim \cite{gordonICCV19splitnet} problem (training on one simulator and then evaluating on another one). 

However, same as in non-photorealistic simulators, when the simulated robot presents changes between training and evaluation (robot-to-robot transfer), its performance is strongly affected. This incapacity to generalize among robots can be addressed using techniques like domain randomization \cite{PengICRA2018}, but such approaches do not constitute a definitive answer, as the overall performance is still reduced. We show evidence of these phenomena in Section \ref{sec:results}, where we also show how our method is transferable among robots, indicating its suitability for the sim-to-real problem.

\subsection{Partial Observability and External Memory}
RL models problems as MDPs, assuming that the information from the environment at a single time step can determine the transition distribution. While table games and some videogames are examples of these problems, most real-world problems, as the ones often found in robotics, are almost all partially observable. In this setting, RNN allows the integration of information from observations trough time. However, other than RNN, there have not been many attempts to solve this issue from a  theoretical perspective. Very specific, yet successful applications that deal with observability are using 2D external representations of the environment (as maps) \cite{chaplot2018active}, \cite{chaplot2020learning} and hardcoding environment-related information or sub-tasks in the observations (usually requiring an oracle, only available in virtual domains) \cite{guss2019minerlcomp}.

We show how the construction of the observation space can help us address the observability problem. Using a formulation of non-fixed dimensionality allows us to integrate temporal information explicitly.

\subsection{Point Cloud Based ML}\label{sec:related-pc}

When considering 3D applications, choosing an appropriate representation is critical. Voxels are the simplest ones, and although they inherit properties from images (regular domains) and can be processed by extensions of the classic 2D convolutions, they present low efficiency from a memory/computational perspective. Another representation are meshes, which have better computational and memory efficiency but are restricted in terms of their topology. Finally, we consider the use of point clouds, and although they are more difficult to process, do not have topological limitations, are memory-efficient, and can be obtained directly from standard sensors (e.g., lidars, depth cameras, etc).

Since point clouds are unordered structures, traditional machine learning operations can not be adopted, and specific types of neural networks must be employed.  An early pioneer was PointNet \cite{pointnet}, which solved the problem of order-invariance by pooling features using invariant-to-order functions (e.g, maximum). Since then, convolution-like approaches that can capture information at different scales have been proposed, such as PointNet++ \cite{pointnet2}, PointCNN \cite{pointcnn}, and ShellNet \cite{shellnetICCV19}. While these architectures can extract multi-scale features, as do convolutions, they usually use small batch sizes due to computational limitations. Among them, ShellNet provides a memory-efficient network compatible with RL algorithms requirements, yet it only extracts features concerning vicinities of certain representative points, without encoding the representative points themselves. In this work, we extend this architecture to be able to encode the representative points' spatial information, which is mandatory for RL problems, unlike in classification and segmentation problems.

Most of the point cloud ML-based research focuses on classification, segmentation, or even flow estimation \cite{flownet} tasks. However, all these instances correspond to the supervised learning setting. More recently \cite{Sarmad_2019_CVPR} combines RL and GANs (Generative Adversarial Networks) to perform shape completion. However, since shape completion is supervised in nature, we do not consider it a true RL task. On robotic-oriented tasks, \cite{graspnet} attains excellent results in manipulation using supervised learning and employs point cloud using arguments similar to ours, and \cite{Levia19} uses RL over point clouds as part of a navigation policy, but uses point clouds only as a 2D representation, leaving most of the complexity of the policy to depth images and convolutions.

In this work, we show that policies can be learned on point clouds using purely RL-based signals, and that in challenging conditions the use of point clouds becomes critical to achieve good performances.

\section{Proposed method}\label{sec:method}

We believe that the biggest challenge for sim-to-real transfers lies in the observation space. Previous methods attempting to solve this problem with no re-training (see Section \ref{sec:related}), forcefully match simulated images to the ones from the real-world, focusing solely on texture and lighting aspects of said images, neglecting to address other concerns such as their geometrical interpretation. As a result, these methods are prone to fail when considering the sim-to-real setting, since the agent/robot itself usually differs between simulation and reality. 

Our proposed method aims to build a canonical space (i.e., a formulation in which observations can be projected in a standard form, independent of the robot and non-essential characteristics of the environment), in which to perform RL training, making correct sim-to-real transfers a natural consequence of the formulation of the observation space.

In \cite{JamesCVPR19}, canonical images are designed, images only be canonical from the point-of-view of textures and lighting, and since images depend on its camera parameters, like its field-of-view and placement on the robot (i.e., camera intrinsics and extrinsics), their choice does not qualify under our criteria of canonical space. Depth images, when available, also fulfill the criteria of \cite{JamesCVPR19} for a canonical space, and has been used extensively for sim-to-real research. However, they suffer from the same limitations of color images, so they do not qualify for our requirements.

Given this, we consider depth sensors such as RGBD or depth images for our observations space, but instead of considering their information as images (i.e., 2D arrays), we consider them as point clouds. Even so, point clouds in the camera reference system $(x_{cam},y_{cam}, depth)$ still depend, as mentioned previously, on its parameters. To circumvent this problem, we project the point clouds to a fixed frame of the robot (e.g., its base, which can be real for wheeled robots, or imaginary for legged ones), obtaining a collection of points $(x_{frame},y_{frame}, z_{frame})$, which are agnostic of the camera itself. Although this method assumes that the camera intrinsics and extrinsics are always known, that is usually the case, given an initial camera calibration procedure and the robot's kinematic chain.

The choice of point clouds over other 3D representations like meshes and shape primitives is not arbitrary. Although other 3D representations also present the desired properties to build a canonical space, their use requires additional estimation and/or processing, whereas point clouds are available directly from common robotic sensors.

We argue that a canonical space that possesses the previously stated requirements, can overcome the problem of sim-to-real. The issues related to sim-to-real, as stated in Section \ref{sec:related}, come from the difference between simulations and reality, which can be roughly separated as differences between environments (sim-to-sim), changes in the observation space (sim-to-sim and robot-to-robot), and differences in the action space (robot-to-robot). It follows that a system successful for sim-to-real, should also be able to succeed in sim-to-sim and robot-to-robot transfers. Even though the use of the canonical space addresses directly the robot-to-robot aspects and partially the sim-to-sim due to a better generalization, we also consider the use of photo-realistic simulators to improve the sim-to-sim performance, and domain-randomization to address the action space concerns of robot-to-robot.

Unlike sim-to-real, robot-to-robot and sim-to-sim, can be evaluated quantitatively in simulations, and we validate those aspects in Section \ref{sec:results}. It should also be noted that while the robot-to-robot is necessary to sim-to-real, it is also a goal in itself as it accomplishes independent-to-robot policies.

One of the biggest challenges several robotic tasks, is working under visually constrained scenarios (e.g., manipulating objects when there are occlusions, or navigating through doors and cluttered environments). This challenge, comes from the partial observability generated by the sensors limitations (e.g., their resolution the case of a lidar, or the field of view in the case of a camera), making critical information not available to the agent, and thus limiting its performance. While this problem can be addressed integrating temporal information of the observations through the use of RNNs, fine-grained actions become difficult due to the stochastic nature of the environments and the difference between simulations and real-world \cite{Levia19}.

To address this issue, since the point clouds have non-fixed dimensionality (unlike images, which are 2D-arrays), we can directly integrate the information from previous time steps so that the agent has access to critical information at all times (e.g., can model walls  and door completely while navigating through them). An additional advantage of using the point cloud space is that it becomes direct to also integrate different depth sensors into a unique representation, avoiding the need for using multi-modal approaches if multiple sensors are available/required. The integration of information from different time steps is not trivial, but it can be achieved using methods such as Self Localization and Mapping (SLAM), visual odometry, flow estimation or point cloud registration. 

\begin{figure*}
  \vspace*{1mm}
  \center
  \includegraphics[width=\textwidth]{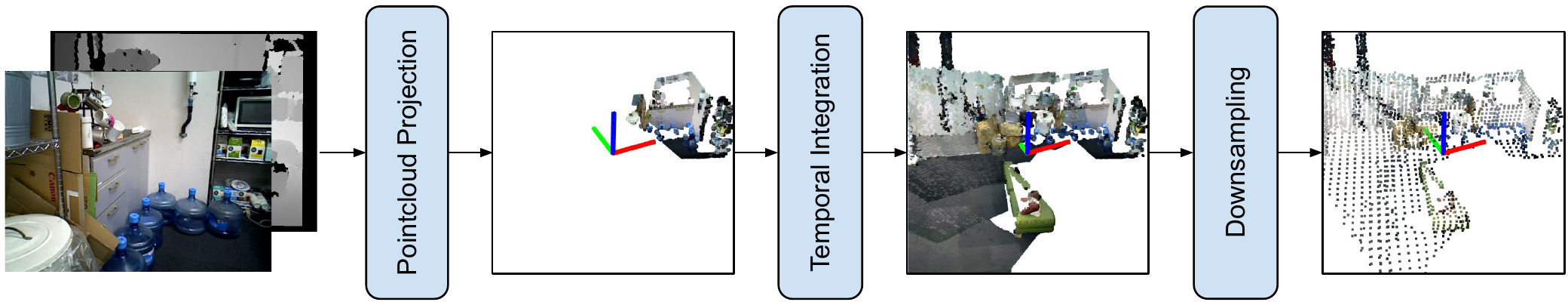}

  \caption{Pipeline of the proposed method. The RGBD image is projected into a robot’s frame (e.g., the robot’s base) using robot specific information, making it highly invariant to the particular robot. Afterward, previous observations are explicitly integrated, addressing partial observability. Finally, the resulting point cloud is down-sampled due to practical considerations.}
  \label{fig:pipeline}
  \vspace*{-1mm}
\end{figure*}

A summary of the proposed method is presented  in Figure \ref{fig:pipeline}. The projection module converts the point cloud from the image space the cartesian space into a fixed frame's coordinate system (e.g., a wheeled robot's base or the center of a drone), by making use of the robot dependent sensor properties (camera intrinsics and extrinsics), and the kinematic chain, with aims to overcome the robot-to-robot problem. Then, the representation goes through a frame integration process, in which point clouds from previous time-steps are registered to the current one to increase performance by improving observability. Finally, the resulting representation is used by a standard RL agent, as explained in Section  \ref{sec:method-navigation}. 

\subsection{Visual Navigation Task}\label{sec:method-navigation}

To validate the proposed methodology, we evaluate it in visual navigation tasks as they provide an ideal case of study to test the hypotheses of Section \ref{sec:method}. Of the several visual navigation-related tasks, we choose \emph{Pointgoal Navigation}, which consists of navigating to a certain target given its coordinates, as it is the most studied task with several previous attempts of sim-to-real \cite{Savva2019}, \cite{LobosTsunekawa2018}, \cite{Levia19}. However, the advantages of the proposed methodology are more beneficial in the case of more complex tasks, like semantic-target navigation, or instruction following.

As mentioned in Section \ref{sec:method}, besides the use of the point cloud canonical space, we use photo-realistic simulators to address the sim-to-sim problem. In this case, we make use of \emph{Habitat} \cite{Savva2019}, which provides a suite for large-scale standardized experiments, with navigation environments reconstructed from real-world scenarios from databases like \emph{MatterPort3D (MP3D)} \cite{Matterport3D} and \emph{Gibson} \cite{gibonenv18}, which have been used previously in sim-to-sim related research \cite{Savva2019}, \cite{gordonICCV19splitnet}.

Additionally, we use domain randomization as part of our approach for robot-to-robot transfer of policies, which in the case of visual navigation takes the form of changing the camera intrinsics/extrinsics at the beginning of each episode (randomizing the observation space), and considering a noisy motion model for the base of the robot (randomizing the action space). The choice of the action space also addresses the robot-to-robot transfer by using discrete actions representing motion primitives (in this case, move forward, turn left, turn right, and stop). This primitives are independent of the robot (unlike velocity commands and low-level control), and can be implemented using open-loop control, odometry, or more complex sensor-based approaches.

\begin{figure*}
 \center
  \includegraphics[width=\textwidth]{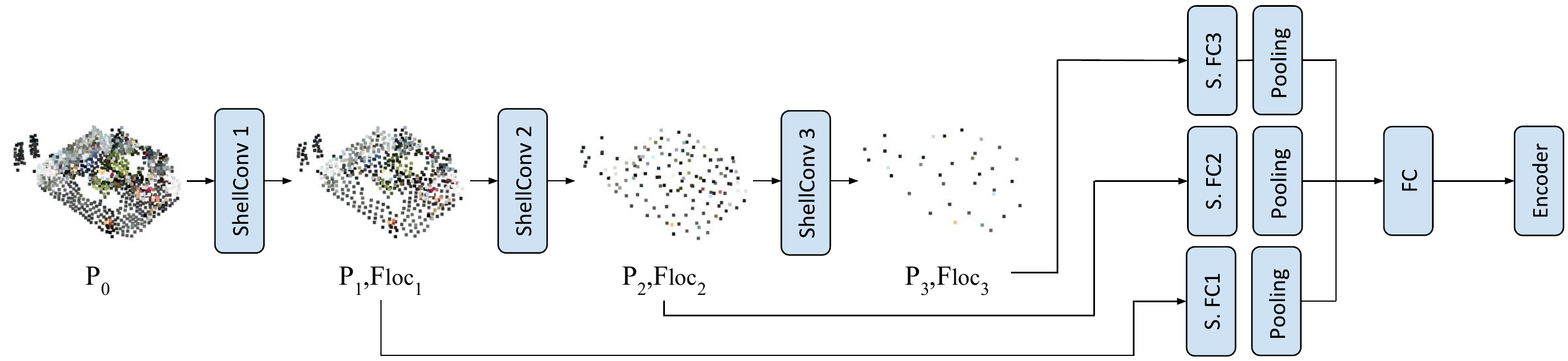}

  \caption{Modified \emph{ShellNet} architecture. $P_{i}$ has dimensions $N_{i}\times 3$, $Floc_{i}$ have dimensions $N_{i}\times f_{i}$, where $N_{i}$ is the decreasing number of points at each layer, and $f_{i}$ is the number of features. Local features do not contain information of their associated points, so that particular information is added using shared-weights fully connected layers as in \cite{pointnet}, and then pooled to obtain a fixed-size vector. Finally, features from different levels are combined using arguments similar to \cite{Szegedy14}.}
  \label{fig:network}
  \vspace*{-2mm}
\end{figure*}

As indicated in Section \ref{sec:related-pc}, the use of point clouds implies the need for point cloud neural networks, and we consider the use of two particular architectures. The first one corresponds to PointNet, since it is computationally the most inexpensive both in terms of memory and computational power, making it suitable for robotic tasks, where real-time operation must be guaranteed. Additionally, we also consider \emph{ShellNet}, which computes multi-resolution features, which can lead to improved performances. The choice of \emph{ShellNet} over other popular convolution-like networks such as \emph{PointNet++} is because it is the only one compatible with the batch size requirements of recent RL algorithms in terms of memory. However, \emph{ShellNet} calculates features concerning vicinities of certain representative points, without encoding the representative points themselves, making it unfit for RL applications, where that geometrical information is mandatory. For this reason, we extend the network architecture as presented in Figure \ref{fig:network}, using a strategy similar to \cite{Szegedy14} to combine information from multiple resolutions, albeit concatenating vecor features instead of feature maps.

\begin{align}
  \mathrm{rew(t)} = \mathrm{-slackness} - \triangle \mathrm{dist_{goal}(t)} + \mathrm{success(t)}\label{eq:rew}
\end{align}

While the formulation of \emph{Habitat} is convenient for AI purposes, when looking for robotic applications, it has several drawbacks. For example, its reward (see Equation \ref{eq:rew}) encourages getting closer to the target (second term), reaching the target (third term, which is implemented as a distance to the target less than a threshold and the stop action), while penalizing stillness using a constant negative reward term dubbed slackness (first term). However, there is no penalty for collisions, resulting in agents that collide almost every episode (see Section \ref{sec:results}). To address this issue, while still being able to compare fairly the reward function with previous research, we do not explicitly modify the reward function, but end training episodes upon a collision (which in turn penalizes collisions as rewards are usually positive in this case). Furthermore, the embodied agent of \emph{Habitat} has a footprint of radius 0.1m, which is true only for a small subset of indoor robots. In this work, we set the footprint radius to 0.25m since it encompasses more indoor robotic platforms (e.g., \emph{HSR}, \emph{Turtlebot}, \emph{Kobuki}, \emph{Pepper}, etc.).

\section{Experimental Results}\label{sec:results}

In this section, we provide experimental results to evaluate the proposed hypotheses of this paper. Although the final objective is sim-to-real, experiments of Section \ref{sec:results-simulated} are performed in simulations to evaluate the robot-to-robot and sim-to-sim capabilities of the proposed method. Then, Section \ref{sec:results-real} presents experiments in a real robot to provide concrete sim-to-real proof.

In addition to environments, Habitat also provides a set of image-based baselines, which we use to compare our methodology. The chosen RGBD-based baseline consists of three convolutional and one fully connected layers to encode the observations, an RNN layer to encode the state, and a final fully connected layer to provide the actor and critic, which are trained using the Proximal Policy Optimization (PPO) \cite{SchulmanWDRK17}. Given this, to compare our method fairly, we replace the visual encoder part of the baseline by 3 shared-weight fully connected layers and a global pooling layer in the case of the \emph{PointNet} architecture, and by 3 \emph{ShellConvs} with their respective shared-weight fully connected layers in the case of the \emph{ShellNet} inspired network (see Figure \ref{fig:network}).

\subsection{Simulated Experiments}\label{sec:results-simulated}

As mentioned in Section \ref{sec:method}, the proposed method requires a registration system to integrate the information from different time steps, like SLAM or visual odometry. Since the observations are rendered from databases, classic methods such as ICP (Iterative Closest Point) provide near-perfect results. In light of this, during training, we use ground-truth information to register point clouds, which in turn reduces training time from weeks to days. The down-sample procedure introduced in Figure \ref{fig:pipeline} consists of the following steps: 1) crop the point cloud to a region of interest suited for the task (in this case, a square of 10m around the robot). 2) sub-sample using voxel sampling to produce a spatially uniform point cloud (some networks, like \cite{pointcnn} and \cite{shellnetICCV19}, work better under this scenario). 3) Further sub-sample the point cloud to a fixed amount of points, using random sampling. While this last step is not strictly necessary, it eases the training procedure as data can be batched. Additionally, this random sampling works similarly to the mechanism of dropout, attempting to address the overfitting of the agent. 
In this work, we set the maximum number of points per point cloud during training to 1024, a value common in the literature, and only use the depth information (discarding the RGB component of the point cloud). Other than the convolutional layers, all other hyper-parameters are taken from \emph{Habitat} \cite{Savva2019}. Experiments consist of 100K updates from 6 workers performing rollouts of length 128 (for a total of 76.8M environment steps). Each particular experiment is repeated 5 times except ones involving \emph{ShellNet}, which are repeated twice, and the reported values correspond to the mean and standard deviation of the models which reported the best rewards in the evaluation set of \emph{MP3D}, within a set of checkpoints taken every 4K updates. We consider the following performance indexes: episode accumulated reward, SPL, and success rate. Experiments are mostly run on NVIDIA’s Titan X GPUs, taking approximately 4 days per configuration, except experiments involving \emph{ShellNets}, which use NVIDIA Quadro RTX due to high memory usage (35GB) and take approximately 2 weeks.

The first experiment aims to demonstrate that high-performance policies can be obtained using only RL-based signals using the proposed method, and justify the change in the reward proposed in Section \ref{sec:method-navigation}. To do this, we compare the \emph{PointNet}-based agent with the RGBD-based agent, and other than the agent's footprint, all remaining aspects, including the training reward, remain the same as in \cite{Savva2019}. Table \ref{tab:exp-basicrew} presents the results of this experiment where can observe that the \emph{PointNet}-based agent obtains performances similar to its RGBD counterpart, both solving practically all evaluation scenes, yet the baseline still outperforms the proposed method slightly in terms of reward and SPL. However, the results change if the same trained policies are evaluated against metrics that consider safety concerns (reward$\star$, SPL$\star$, and collision rate$\star$), that is, considering collisions as failed episodes and thus ending them prematurely. Table \ref{tab:exp-basicrew} shows how in both cases, the overall performance gets close to zero, making them unfit for real applications.

\begin{table}[]
\vspace*{2mm}
\centering
\caption{Comparison under conditions presented in \cite{Savva2019}}
\label{tab:exp-basicrew}
\begin{tabular}{l|l|l|}
\cline{2-3}
                                    & RGBD & PointNet \\ \hline
\multicolumn{1}{|l|}{Reward}        & $\mathbf{18.16 \pm 0.03}$   & $18.01 \pm 0.06$        \\ \hline
\multicolumn{1}{|l|}{SPL}           & $\mathbf{0.97 \pm 0.01}$   & $0.93 \pm 0.01$        \\ \hline
\multicolumn{1}{|l|}{Success rate}  & $\mathbf{0.99 \pm 0.00}$   & $\mathbf{0.99 \pm 0.00}$        \\ \hline
\multicolumn{1}{|l|}{Reward$\star$}       & $1.03 \pm 0.03$  & $\mathbf{1.19 \pm 0.10}$        \\ \hline
\multicolumn{1}{|l|}{SPL$\star$}          & $0.02 \pm 0.01$   & $\mathbf{0.05 \pm 0.01}$        \\ \hline
\multicolumn{1}{|l|}{Success rate$\star$} & $0.02 \pm 0.01$   & $\mathbf{0.05 \pm 0.01}$       \\ \hline
\end{tabular}
\vspace{-4mm}
\end{table}

To evaluate if the agents can succeed when considering safety concerns (i.e., considering collisions as failures), the experiments are repeated using reward$\star$ during training (from here on, all experiments use reward$\star$ and so the $\star$ notation is dropped). The results of this experiment are presented in Table \ref{tab:collisionrew} (first rows), where it can be observed that under these new conditions, both agents achieve good performances.

A second set of experiments evaluates the capability of the proposed method to address the robot-to-robot setting, by using both domain randomization as presented in Section \ref{sec:method-navigation} during training and evaluation, and the point cloud canonical space from Section \ref{sec:method}. Table \ref{tab:envrandom} presents the corresponding results, which confirm the superior performance of the point cloud-based agent, validating the proposed method in terms of robot-to-robot transfer. In this experiment, we also compare \emph{PointNet} against \emph{ShellNet}, and Table \ref{tab:envrandom} indicates that even though \emph{ShellNet} has advantages over \emph{PointNet} (see Section \ref{sec:related-pc}), for the \emph{Point Goal} task and the used number of points, both networks achieve very similar results.

\begin{table}[]
\vspace*{2mm}
\centering
\caption{Robot-to-robot evaluation under domain randomization as presented in Section \ref{sec:method-navigation}}
\label{tab:envrandom}
\begin{tabular}{l|l|l|l|}
\cline{2-4}
                                   & RGBD & PointNet & ShellNet \\ \hline
\multicolumn{1}{|l|}{Reward}       & $9.21 \pm 0.27$    & $\mathbf{13.91 \pm 0.16}$        & $13.21 \pm 0.23$        \\ \hline
\multicolumn{1}{|l|}{SPL}          & $0.44 \pm 0.02$    & $\mathbf{0.68 \pm 0.02}$        & $\mathbf{0.68 \pm 0.01}$        \\ \hline
\multicolumn{1}{|l|}{Success Rate} & $0.54 \pm 0.02$    & $\mathbf{0.82 \pm 0.02}$        & $0.81 \pm 0.01$        \\ \hline
\end{tabular}
\end{table}

An additional ablation study is performed, to identify the impact of the canonical space (RGBD vs. point cloud) under domain randomization (both in training and evaluation). Condition A of Table \ref{tab:collisionrew} involves changing the camera parameters, whereas Condition B involves the noisy action model presented in Section \ref{sec:method-navigation}. Results indicate that when faced with different sensor configurations than the ones used in training, the baseline has near-null performance  since the changes in the sensor make the learned geometric interpretation invalid, whereas the proposed method barely has its performance affected since sensor changes are modeled in the representation, validating the idea of it being a canonical space. However, in Condition B, both agents have their performances reduced, which indicate that neither agent is robust against changes in the action space, and that environment randomization is still needed.

\begin{table}[]
\centering
\caption{Robot-to-robot ablation study, considering the effects of the different types of domain randomization and the canonical space}
\label{tab:collisionrew}
\begin{tabular}{ll|l|l|}
\cline{3-4}
                                                           &              & RGBD & PointNet \\ \hline
\multicolumn{1}{|l|}{\multirow{3}{*}{Basic Eval}}          & Reward       & $12.95 \pm 0.25$   & $\mathbf{14.64 \pm 0.24}$  \\ \cline{2-4} 
\multicolumn{1}{|l|}{}                                     & SPL          & $\mathbf{0.81 \pm 0.04}
$ & $0.78 \pm 0.03
$ \\ \cline{2-4} 
\multicolumn{1}{|l|}{}                                     & Success rate & $\mathbf{0.87 \pm 0.03}$ & $\mathbf{0.87 \pm 0.02}$ \\ \hline
\multicolumn{1}{|l|}{\multirow{3}{*}{\begin{tabular}[c]{@{}l@{}}Eval conditions\\ A\end{tabular}}}   & Reward       & $0.53 \pm 0.17$ & $\mathbf{14.05 \pm 0.24}$ \\ \cline{2-4} 
\multicolumn{1}{|l|}{}                                     & SPL          & $0.08 \pm 0.01$ &  $\mathbf{0.75 \pm 0.04}$         \\ \cline{2-4} 
\multicolumn{1}{|l|}{}                                     & Success rate & $0.10 \pm 0.01$ & $\mathbf{0.85 \pm 0.02}$ \\ \hline
\multicolumn{1}{|l|}{\multirow{3}{*}{\begin{tabular}[c]{@{}l@{}}Eval conditions\\ B\end{tabular}}}   & Reward       & $9.89 \pm 0.50$ & $\mathbf{12.00 \pm 0.34}$ \\ \cline{2-4} 
\multicolumn{1}{|l|}{}                                     & SPL          & $0.57 \pm 0.02$ & $\mathbf{0.61 \pm 0.00}$\\ \cline{2-4} 
\multicolumn{1}{|l|}{}                                     & Success rate & $0.65 \pm 0.03$ & $\mathbf{0.73 \pm 0.01}$ \\ \hline
\multicolumn{1}{|l|}{\multirow{3}{*}{\begin{tabular}[c]{@{}l@{}}Eval conditions\\ A+B\end{tabular}}} & Reward       & $0.73 \pm 0.23$ & $\mathbf{11.25 \pm 0.27}$ \\ \cline{2-4} 
\multicolumn{1}{|l|}{}                                     & SPL          & $0.06 \pm 0.01$ &  $\mathbf{0.56 \pm 0.02
}$\\ \cline{2-4} 
\multicolumn{1}{|l|}{}                                     & Success rate & $0.08 \pm 0.01$ & $\mathbf{0.69 \pm 0.01}$ \\ \hline
\end{tabular}
\vspace{-3mm}
\end{table}

The final experiment evaluates the sim-to-sim scenario. For this, we train and evaluate the agents in \emph{Gibson} besides \emph{MP3D}, and consider as a performance measure the decrease of performance produced by training the agents in a different simulator than the ones they are evaluated in. The results from Tables \ref{tab:sim2sim-rgbd} and \ref{tab:sim2sim-pointcloud} indicate that for the \emph{MP3D}$\to$\emph{Gibson} transfer, both agents obtain near-perfect transfer (training in \emph{MP3D} obtains the same result when evaluating in \emph{Gibson} than when the policy was also trained in \emph{Gibson}). When observing the \emph{Gibson}$\to$\emph{MP3D} case, however, it can be observed that the performance of the point cloud agent was not lowered as much as that of its RGBD counterpart, evidencing that the proposed method has a higher generalization (higher sim-to-sim performance) when transferred to more challenging scenarios (as \emph{MP3D} environments being more difficult than \emph{Gibson} ones \cite{Savva2019}, \cite{gordonICCV19splitnet}).

\begin{table}[]
\vspace*{2mm}
\centering
\caption{Sim-to-sim results for the RGBD-based agent}
\label{tab:sim2sim-rgbd}
\begin{tabular}{ll|l|l|}
\cline{3-4}
                       &              & Train MP3D & Train Gibson \\ \hline
\multicolumn{1}{|l|}{\multirow{3}{*}{\begin{tabular}[c]{@{}l@{}}Eval\\ MP3D\end{tabular}}}   & Reward & $9.21 \pm 0.27$ & $2.06 \pm 0.35$ \\ \cline{2-4} 
\multicolumn{1}{|l|}{} & SPL          & $0.44 \pm 0.02$         & $0.11 \pm 0.02$           \\ \cline{2-4} 
\multicolumn{1}{|l|}{} & Success Rate & $0.54 \pm 0.02$         & $0.15 \pm 0.03$           \\ \hline
\multicolumn{1}{|l|}{\multirow{3}{*}{\begin{tabular}[c]{@{}l@{}}Eval\\ Gibson\end{tabular}}} & Reward & $4.44 \pm 0.35$ & $4.55 \pm 0.21$ \\ \cline{2-4} 
\multicolumn{1}{|l|}{} & SPL          & $0.27 \pm 0.01$         & $0.28 \pm 0.01$           \\ \cline{2-4} 
\multicolumn{1}{|l|}{} & Success Rate & $0.55 \pm 0.01$        & $0.40 \pm 0.01 $          \\ \hline
\end{tabular}
\end{table}

\begin{table}[]
\centering
\caption{Sim-to-sim results for the point cloud-based agent}
\label{tab:sim2sim-pointcloud}
\begin{tabular}{ll|l|l|}
\cline{3-4}
                       &              & Train MP3D & Train Gibson \\ \hline
\multicolumn{1}{|l|}{\multirow{3}{*}{\begin{tabular}[c]{@{}l@{}}Eval\\ MP3D\end{tabular}}}   & Reward & $13.91 \pm 0.16$ & $9.51 \pm 0.55$ \\ \cline{2-4} 
\multicolumn{1}{|l|}{} & SPL          & $0.68 \pm 0.02$         & $0.45 \pm 0.02$           \\ \cline{2-4} 
\multicolumn{1}{|l|}{} & Success Rate & $0.82 \pm 0.02$         & $0.64 \pm 0.03$           \\ \hline
\multicolumn{1}{|l|}{\multirow{3}{*}{\begin{tabular}[c]{@{}l@{}}Eval\\ Gibson\end{tabular}}} & Reward & $9.16 \pm 0.33$ & $9.52 \pm 0.23$ \\ \cline{2-4} 
\multicolumn{1}{|l|}{} & SPL          & $0.54 \pm 0.01$         & $0.56 \pm 0.01$           \\ \cline{2-4} 
\multicolumn{1}{|l|}{} & Success Rate & $0.73 \pm 0.01$        & $0.74 \pm 0.01$          \\ \hline
\end{tabular}
\vspace{-5mm}
\end{table}

\subsection{Real-world Experiments}\label{sec:results-real}

To evaluate the sim-to-real capabilities of the proposed method (with no fine-tuning of the policy), experiments are performed in a Toyota’s HSR platform, which is a traditional service robot, equipped with depth sensors, and an articulated head capable of moving in the z-axis, beside yaw and tilt head movement (this also enable us to test robot-to-tobot capabilities by changing these parameters in each episode). It is important to note that the characteristics of the HSR did not influence in any case the design choices or the experiments of Section \ref{sec:results-simulated}, so it can be considered a fair and unbiased sim-to-real evaluation. Perhaps the only exception consists of the base footprint, which as mentioned before, was chosen to be general to most robotic platforms. \\

The interface with the robot and the related systems required is implemented through \emph{ROS (Robot Operative System)}, which is the most commonly used robot software platform. As mentioned in Section \ref{sec:method}, the proposed method requires a registration system to integrate observations through time. Although in simulations this is done using ground-truth information as explained in Section \ref{sec:results-simulated}, in this real-world validation, we make use of \emph{ORB-SLAM2} \cite{ORBSLAM2} as it is known to provide good out-of-the-box performance. Although it is possible to process the point clouds frame-by-frame, we instead only integrate the point clouds associated with \emph{ORB-SLAM2} \emph{keyframes}, which present a good temporal/space resolution (Figure \ref{fig:pipeline} presents examples of the point clouds obtained through this process). Although point cloud processing is known to be computation-intensive, by processing only the latest \emph{keyframes}, the overhead of the system, which runs as part of ORB-SLAM, is less than 10ms, making it suitable for robot applications.

In addition to the point cloud integration system, the kinematic chain is obtained from the  \emph{HSR} magnetic sensors, the agent's action space is implemented through a simple differential controller over the base, and the \emph{Point Goal} target is updated through the localization provided by \emph{ORB-SLAM2}. Since the agents' policies use small neural networks, they can be executed with no problem using the onboard NVIDIA Jetson and Intel Intel i7-4700EQ in real-time (having an inference time of 30ms for the image-based agent and 60ms for the point cloud version in the on-board CPU). All the experiments in this Section are carried in the facilities of the Machine Intelligence Laboratory of the University of Tokyo.

\begin{figure}
   \vspace*{2mm}
    \centering
    \begin{subfigure}[b]{0.22\textwidth}
        \centering
        \captionsetup{justification=centering}
        \includegraphics[width=\textwidth]{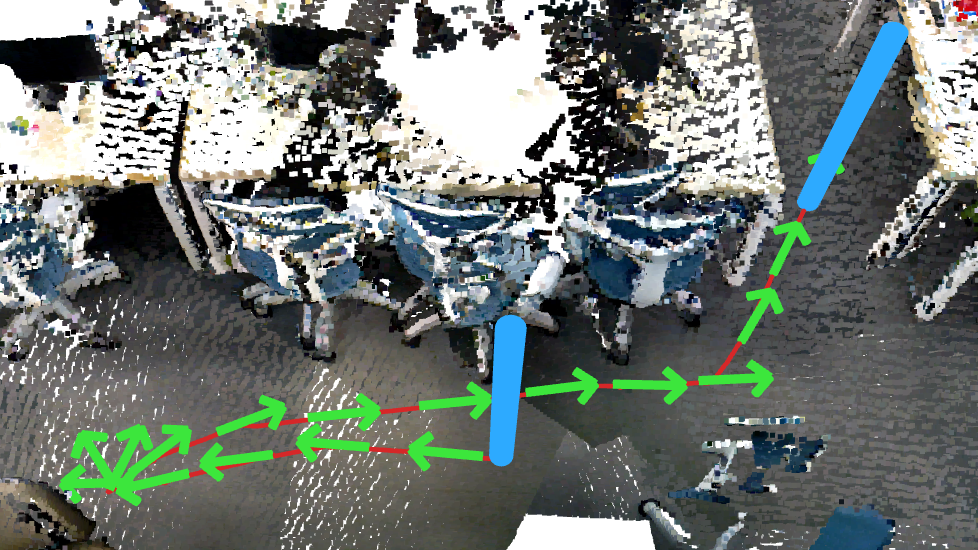}
        \caption{Sub optimal trajectory}
        \label{fig:images-simulated}
    \end{subfigure}
    \quad
    \begin{subfigure}[b]{0.22\textwidth}
        \centering
        \captionsetup{justification=centering}
        \includegraphics[width=\textwidth]{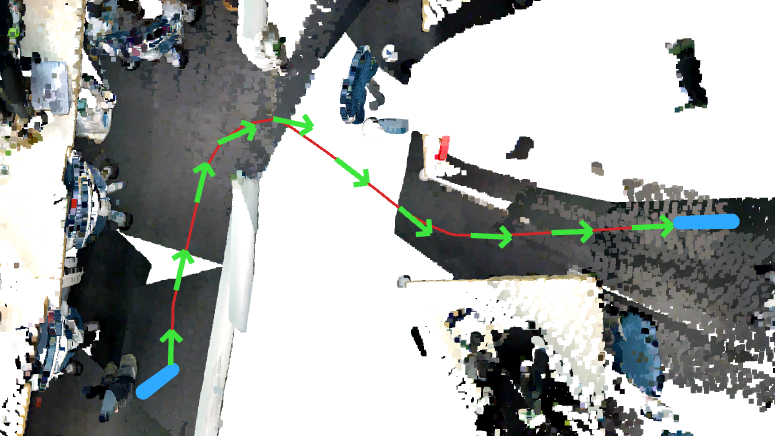}
        \caption{Successful episode}
        \label{fig:images-nao}
    \end{subfigure}
    \\
    \vspace*{2.5mm}
    \begin{subfigure}[b]{0.22\textwidth}
        \centering
        \captionsetup{justification=centering}
        \includegraphics[width=\textwidth]{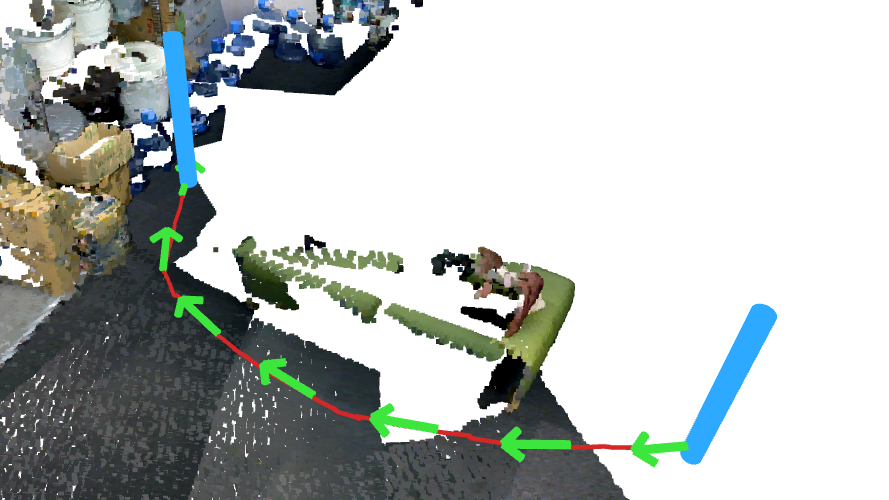}
        \caption{Successful episode}
        \label{fig:images-simulated-segmented}
    \end{subfigure}
    \quad
    \begin{subfigure}[b]{0.22\textwidth}
        \centering
        \captionsetup{justification=centering}
        \includegraphics[width=\textwidth]{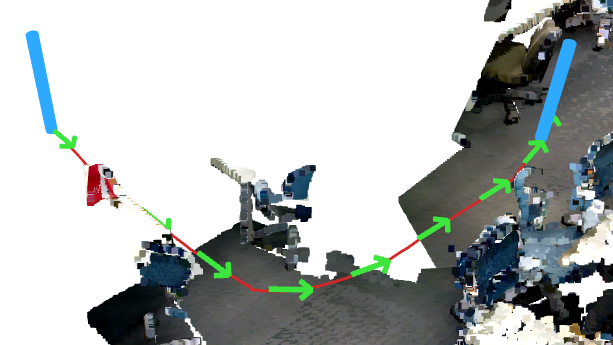}
        \caption{Successful episode}
        \label{fig:images-nao-segmented}
    \end{subfigure}
    \caption{Examples of navigation trajectories in real-world experiments.}
    \label{fig:trajectories}
    \vspace*{-5mm}
\end{figure}

Experiments performed in the HSR use the complete method, including domain randomization during training in simulation, and mimic the domain randomization during the real-world evaluation by moving the head of the robot, including its height to new positions before each task. While the RGBD-baseline can solve simple tasks (e.g., obstacle-less paths), it fails in most non-trivial cases. On the other hand, the proposed agent can navigate through most configurations. Some examples of agent trajectories can be observed in Figure \ref{fig:trajectories}, and during experiments, most sub-optimal trajectories occur when targets are behind the robot, and most failures consist on collisions with objects which were never in the visual range, and collisions with the edge of the robot against difficult objects like chair legs. On the experiments, the agent solves approximately about $75\%$ of the tasks, but the SPL and reward statistics are not calculated due to the lack of a ground-truth system. More examples of trajectories can be observed at \url{https://youtu.be/IJPp9JxvR_c}.

\section{Conclusions and Future Work}\label{sec:conclusions}


In this paper, we proposed an approach for seamless sim-to-real transfer for a visual navigation task based on the designing of a canonical observation space using point clouds, employing domain randomization, and using photo-realistic simulators. Simulated results
 show that the proposed method is largely unaffected by unseen configurations in the robot-to-robot setting, whereas image-based baselines fail completely, and that the proposed method achieves better generalization than the baseline in the sim-to-sim setting when the target simulator is more complex than the one used in training. During real-world experiments, only the proposed method presents a correct sim-to-real performance in an out-of-the-box fashion. However, several limitations are identified in this work. Upon careful observation of the failed episodes, most corresponds to collisions with objects which were never seen directly, indicating that RL signals alone may not be enough and that self-supervised curiosity and common sense could be considered in the point cloud space. Additionally, the proposed method requires an external point cloud-registration system, which limits the performance and potential applications of the method. Furthermore, due to computational limitations, point cloud-based neural networks need to use a relatively small number of points, and all the benefits of modern architectures could not be used to their full potential. Finally, although experiments were shown for a particular task for the sake of comparison with previous research, it is expected that this method presents more benefits for tasks like active mapping, active vision, and more challenging visual navigation tasks.

\bibliographystyle{./IEEEtran} 
\bibliography{./IEEEabrv,./mendeley}
\end{document}